\documentclass[twoside]{article}

\usepackage[accepted]{aistats2021}
\usepackage{amsmath}
\usepackage{amsfonts}
\usepackage{amsthm}
\usepackage{booktabs}
\usepackage{multirow}

\usepackage{natbib}
\usepackage{hyperref}

\usepackage{graphicx}
\usepackage{subfigure}

\usepackage{color}

\newtheorem{Definition}{Definition}
\newtheorem{Assumption}{Assumption}
\newtheorem{Theorem}{Theorem}
\newtheorem{Lemma}{Lemma}

\newcommand\blfootnote[1]{%
  \begingroup
  \renewcommand\thefootnote{}\footnote{#1}%
  \addtocounter{footnote}{-1}%
  \endgroup
}

\begin{document}

%

%
\runningauthor{Hejia Qiu, Chao Li, Ying Weng, Zhun Sun, Xingyu He, Qibin Zhao}

\twocolumn[

\aistatstitle{On the Memory Mechanism of Tensor-Power Recurrent Models
}

\aistatsauthor{ Hejia~Qiu$^\star$ \And Chao~Li$^{\star,\S}$}

\aistatsaddress{ University of Nottingham Ningbo China\\University of Nottingham UK\And RIKEN-AIP\\Tokyo, Japan}



\aistatsauthor{ Ying~Weng$^{\S}$ \And Zhun~Sun}

\aistatsaddress{ University of Nottingham Ningbo China\\University of Nottingham UK \And BIGO PTE. LTD.\\Singapore} 

\aistatsauthor{Xingyu~He \And Qibin~Zhao$^{\S}$}

\aistatsaddress{University of Nottingham Ningbo China\\University of Nottingham UK \And RIKEN-AIP\\Tokyo, Japan} 

]


\begin{abstract}
Tensor-power~(TP) recurrent model is a family of non-linear dynamical systems,
of which the recurrence relation consists of a $p$-fold~(\textit{a.k.a.}, degree-$p$) tensor product.
Despite such the model frequently appears in the advanced recurrent neural networks~(RNNs), to this date there is limited study on its memory property, a critical characteristic in sequence tasks.
In this work, we conduct a thorough investigation of the memory mechanism of TP recurrent models.
Theoretically, we prove that a large degree $p$ is an essential condition to achieve the long memory effect, yet it would lead to unstable dynamical behaviors.
Empirically, we tackle this issue by extending the degree $p$ from discrete to a differentiable domain, such that it is efficiently learnable from a variety of datasets.
Taken together, the new model is expected to benefit from the long memory effect in a stable manner. 
We experimentally show that the proposed model achieves competitive performance compared to various advanced RNNs in both the single-cell and seq2seq architectures.

\end{abstract}

\section{Introduction}\label{sec:intro}

Recurrent neural networks~(RNNs) have been popularly used for tasks arising in domains such as time series analysis and natural language processing.
They are powerful because the recurrent dynamics of the hidden states allow the models to remember the past information.
\blfootnote{$^\star$Equal Contribution.}\blfootnote{$^{\S}$Correspondence to: Chao~Li$<$\href{mailto:chao.li@riken.jp}{chao.li@riken.jp}$>$;\\Ying~Weng$<$\href{mailto:yingweng@gmail.com}{yingweng@gmail.com}$>$;
\\
Qibin~Zhao$<$\href{mailto:qibin.zhao@riken.jp}{qibin.zhao@riken.jp}$>$}
In the standard RNNs, the transition function of the hidden states obeys a linear transform\footnote{It is an affine transform more precisely.}
following an element-wise activation function, and the vanilla form of RNN has an inherent difficulty to learn the long-range dependence of the data~\citep{bengio1994learning}.
As alternatives, a family of variants of RNNs are proposed, where the linear form of the transition function is extended to higher-degree polynomials~\citep{sutskever2011generating,wu2016multiplicative,schlag2018learning}. 
Rich results by numerical experiments demonstrate the RNNs equipped with polynomial transitions are more expressive than their vanilla counterparts~\citep{yu2017long,su2020convolutional}.

However, those polynomial-induced variants fail to become mainstream models in sequence tasks.
The cause is mainly twofold:
theoretically, it remains unclear why the higher-degree models outperform their linear counterparts;
empirically, an ``over-large'' degree would lead to the model being unstable in both the training and inference phases.
Moreover, the model size would explode when the degree increases.
Although for the latter it is alleviated by tensor decomposition~(TD)~\citep{ye2018learning,pan2019compressing}, 
the selection of the degree parameters is still achieved with the inefficient exhausitive search~\citep{yu2017learning,hou2019deep}.
As a consequence, the state of affairs raises important unresolved questions.
\emph{How does the model degree determine the memory property and dynamical behaviors?}
\emph{Is it possible to efficiently select the optimal degree in practice?}

The goal of this work is to shed light on both two questions through a theoretical and empirical investigation.
We focus on a unified expression of the aforementioned variants of RNNs named \emph{tensor-power~(TP) recurrent models}, in which the degree-$p$ polynomial is reformulated as a $p$-fold tensor product of the hidden states multiplied by a weight tensor.

Theoretically, we prove the long memory property of the model requires a large value of the degree $p$~\mbox{(see \emph{Thm. \ref{thm:main}})}, while in this case the unstable behaviors of the model are inherently inevitable~(see \emph{Thm.~\ref{thm:unbound}}).
Empirically, we propose a degree-learnable method to seek a ``saddle-point'' of balancing the long memory effect and stability of the model.
In particular, we extend the feasible range of the model degree from discrete to a continuous domain using tensor decomposition~(TD).
The differentiable essence of the degree allows us to efficiently learn it from datasets by stochastic gradient descent~(SGD) and its variants.
Endowing with the additional freedom on the degree parameter, we expect that the new model can benefit from the long memory effect yet keep the stability reachable.
Extensive numerical results demonstrate the superior performance of the proposed model in time series forecasting tasks with both shallow and relatively deep architectures.

\subsection{Related works}

\noindent\textbf{Polynomial recurrent neural networks~(RNNs)}.
The earliest study of recurrent neural networks~(RNNs) with polynomial transition dates back to literature~\citep{lee1986machine,giles1990higher,pollack1991induction}.
Later, similar architectures are used in various machine learning tasks~\citep{sutskever2011generating,wu2016multiplicative}.
More recently, the connection between higher-degree polynomial RNNs and the weighted automata~(WA) is uncovered through spectral learning~\citep{rabusseau2019connecting}, and on the practical side the polynomials are also applied to constructing the correlation among time-steps in multimodal tasks~\citep{hou2019deep,litpfn}.
Compared to those works, we are the first to theoretically investigate the memory property of the model, and the proposed method has the capability of learning the optimal degree parameters from datasets while the parameters are assumed to be fixed in previous works.




\noindent\textbf{Tensor methods in RNNs.}
There are three lines of studying the tensor methods in RNNs.
One line of work is the extension of the polynomial RNNs to higher-degree.
In the existing works, the most closed one to ours is HOTRNN~\citep{yu2017long,yu2017learning}, 
and the similar idea is extended in recent works on convolutional LSTM~\citep{huang2020physics,su2020convolutional}.
Compared to those methods, we mainly focus on the issue how the ``tensor order''~(\textit{i.e.}, the ``degree'' in this work) influences the performance of RNNs and how to obtain the optimal tensor order in a more efficient manner rather than by the exhaustive search.
The second line is to apply tensor decomposition~(TD) to compressing RNNs~\citep{tjandra2017compressing,yang2017tensor,ye2018learning,mehta2019scaling,pan2019compressing,wang2020kronecker}.
Unlike ours, they still model the transition function as a ``linear'' transform, yet reshaping the weights into high-order tensors.
The third line of work is to use the tensor network modeling to analyze the expressive power of RNNs~\citep{khrulkov2017expressive,khrulkov2019generalized}.
In contrast, our work pays more effort on the memory property and focuses on the impact by the ``tensor-order'' rather than different decomposition models.


\noindent\textbf{RNNs for long memory.}
Many works have been proposed on the long memory property of RNNs, for instance, the works by~\cite{le2018learning,trinh2018learning,voelker2019legendre,wang2019state,lechner2020learning} to name a few.
In this work, the theoretical aspect is partially inspired by recent studies~\citep{greaves2019statistical,zhao2020rnn} from the stochastic perspective and the works by~\cite{pascanu2013difficulty,miller2018stable} from the stability perspective.
Compared to those works, we focus on more specific non-linear dynamics of the transition and analysing how the model degree influences the memory property.

The study on multivariate factional polynomial fitting originates by~\cite{royston1994regression} in statistics, and is also discussed in the deep learning community~\citep{gulcehre2014learned,sun2018feature}.
Our work on the degree-learnable approach is connected to those works, but focuses on the different issue.
Most recently, we also extend the similar idea into a more general form named fractional tensor network~\citep{li2020high}, which focuses on different tensor structures and feedforward learning models. 

\emph{The proofs and additional experimental details can be found in the supplementary material. Our code is available at
    \url{https://github.com/minogame/AISTATS_2021}.}
    
\section{Preliminaries}
In this section, we first present basic notations of tensor algebra.
After that, we recall the definition of the long memory in statistics in terms of \emph{recurrent network process~(RNP)}~\citep{zhao2020rnn}.
Last, we introduce the \emph{tensor-power~(TP) recurrent model}.

\noindent\textbf{Notation.}
We define a \emph{tensor} as a multi-dimensional array of real numbers~\citep{kolda2009tensor}.
To distinguish from the notion of ``order'' in time-series analysis, we use the term \emph{degree of a tensor} to denote the number of indices.
In the rest of the paper, we use italic letters to denote scalars, \textit{e.g.}, $n,N\in{}\mathbb{R}$, and use boldface letters to denote vectors and matrices, \textit{e.g.}, $\mathbf{h,y}\in{}\mathbb{R}^{n}$ and $\mathbf{W}\in\mathbb{R}^{n\times{}n}$. 
For higher-degree tensors, we denote them by calligraphic letters, \textit{e.g.}, $\mathcal{A,B}\in{}\mathbb{R}^{I_1\times{}I_2\times\cdots\times{}I_d}$.
Sometimes we also use the calligraphic letters to represent vectors or matrices without ambiguity.
The \emph{symmetry} of a tensor is defined as the invariance under arbitrarily reshuffling a sub-collection of  the tensor's indices, and the \emph{index-shift} operation of a tensor is defined as rotating permuting the indices of a tensor in counterclockwise order.
Examples about the  ``symmetry'' and ``index-shifting'' are given in the supplementary material.
For any integer $k$, we use $[k]$ to denote the set of integers from $1$ to $k$.
For any real number $x$, we use $\vert{}x\vert$ and $sgn(x)$ to denote its absolute value and sign, respectively.
If $\mathbf{x}\in{}\mathbb{R}^{I_x}$ and $\mathbf{y}\in{}\mathbb{R}^{I_y}$, we use $\left<\mathbf{x},\mathbf{y}\right>\in{}\mathbb{R}$ and $\mathbf{x}\otimes{}\mathbf{y}\in{}\mathbb{R}^{I_x\times{}I_y}$ to respectively denote the inner and tensor product between vectors, and their straightforward extension to matrices and tensors.
Moreover, we use $\mathbf{x}^{\otimes{}p}\in{}\mathbb{R}^{I_x^p}$ to denote the degree-$p$ \emph{tensor power}~(TP), which represents the $p$-fold tensor product of $\mathbf{x}$ with itself.
For a degree-$q$ tensor $\mathcal{G}\in\mathbb{R}^{n^q}$ and a vector $\mathbf{x}\in{}\mathbb{R}^{n}$, we use $\mathcal{G}\times_i\mathbf{x},\,i\in{}[q]$ to denote \emph{tensor-vector} product along the $i$th index~\citep{cichocki2007nonnegative}.
The spectral norm of a tensor $\mathcal{G}\in{}\mathbb{R}^{I_1\times\cdots\times{}I_q}$ is denoted by $\Vert\mathcal{G}\Vert_2$, which is defined as
\begin{equation}
\begin{split}
    \Vert\mathcal{G}\Vert_{2}
    =&\sup_{\mathbf{u}_i\in{}\mathbb{R}^{I_i},i\in{}[q]}\Vert{}\mathcal{G}\times_1{}\mathbf{u}_1\times_2{}\cdots\times_q\mathbf{u}_q\Vert_2,\\
    &\quad{}\quad{}\quad{}s.t.\,\Vert{}\mathbf{u}_i\Vert_2\leq{}1,\forall{}i
\end{split},\label{eq:TensorOpNorm}
\end{equation}
where $\Vert{}\mathbf{u}_i\Vert_2$ denotes the Euclidean norm of $\mathbf{u}_i$.

\subsection{Recurrent network process~(RNP)}

The notion of \emph{recurrent network process~(RNP)} is defined to analyze the memory mechanism of recurrent neural networks~(RNNs).
Specifically, assume a general RNN with input $\{\mathbf{x}^{(t)}\}$, output $\{\mathbf{y}^{(t)}\}$ and the hidden states $\{\mathbf{h}^{(t)}\}$ where $\mathbf{y}^{(t)}\in{}\mathbb{R}^l$ and $\mathbf{h}^{(t)}\in{}\mathbb{R}^m$, then its RNP is defined as a homogeneous Markov chain with a transition function $M$:
\begin{equation}
    \begin{pmatrix}
    \mathbf{y}^{(t)}\\
    \mathbf{h}^{(t)}
    \end{pmatrix}
    =M
    \begin{pmatrix}
    \mathbf{y}^{(t-1)}\\
    \mathbf{h}^{(t-1)}
    \end{pmatrix}
    +
    \begin{pmatrix}
    \mathbf{\varepsilon}^{(t)}\\
    0
    \end{pmatrix},\label{eq:RNP}
\end{equation}
where $\{\varepsilon^{(t)}\}$ represents a sequence of independent and identically distributed (\textit{i.i.d.}) random vectors.
Eq.~\eqref{eq:RNP} can be used to describe various RNNs like the vanilla RNN and LSTM~\citep{hochreiter1997long} without exogenous inputs.
The term \emph{short or long memory} of the stochastic process generated by Eq.~\eqref{eq:RNP} is roughly defined as below.
More precise definition is given in works by 
~\cite{beran2016long,greaves2019statistical,zhao2020rnn}.
Roughly speaking,
\begin{Definition}[memory of RNPs, roughly]
The process $\left\{\mathbf{s}^{(t)}:=\left(\mathbf{y}^{(t),\top},\mathbf{h}^{(t),\top}\right)^\top\in{}\mathbb{R}^{l+m}\right\}$ generated by Eq.~\eqref{eq:RNP} has short memory if its autocovariance function is summable.
Otherwise, the process has long memory.
\label{thm:Def}
\end{Definition}
The relationship between the memory property and the operator norm of the transition function $M$ is revealed as below:
\begin{Assumption}
\emph{(1)} The joint density function of $\mathbf{\varepsilon}(t)$ is continuous and positive everywhere; \emph{(2)} for some $\kappa\geq{}2$, $\mathbb{E}\Vert{}\mathbf{\varepsilon}^{(t)}\Vert_2^\kappa<\infty$.\label{Thm:Assp}
\end{Assumption}
\begin{Theorem}[by~\cite{zhao2020rnn}]
Under Assumption 1, if there exist real numbers $0<a<1$ and $b$ such that $\Vert{}M(\mathbf{x})\Vert_2\leq{}a\Vert{}\mathbf{x}\Vert_2+b$, then the RNP~\eqref{eq:RNP} has short memory.\label{thm:zhao}
\end{Theorem}


Def. ~\ref{thm:Def} and Thm.~\ref{thm:zhao} will be used in Sec. 3.1 to model and analyze the memory property of the TP recurrent model.
We use them to bridge the gap from the memory property to the tensor power operation.

\subsection{Tensor-power~(TP) recurrent model}
A \emph{degree-}$p$ tensor-power~(TP) recurrent model, with its input $\mathbf{x}^{(t)}\in\mathbb{R}^{l}$ and hidden state $\mathbf{h}^{(t)}\in\mathbb{R}^{m}$ at the time-step $t$, is defined as\footnote{Here the bias term is omitted for brevity.}
\begin{equation}
    \begin{split}
        \mathbf{h}^{(t)} &= \mathcal{G}\times_1
        \begin{pmatrix}
        \mathbf{x}^{(t)}\\
        \mathbf{h}^{(t-1)}
        \end{pmatrix}
        \times_2\cdots\times_p
        \begin{pmatrix}
        \mathbf{x}^{(t)}\\
        \mathbf{h}^{(t-1)}
        \end{pmatrix}\\
        &=\mathcal{G}\cdot
        \begin{pmatrix}
        \mathbf{x}^{(t)}\\
        \mathbf{h}^{(t-1)}
        \end{pmatrix}^{\otimes{}p}
    \end{split},\label{eq:TPRM}
\end{equation}
where $\mathcal{G}\in\mathbb{R}^{n^p\times{}m},\,n=l+m$ denotes a degree-$(p+1)$ tensor, which is partially symmetric involving the first $p$ indices. The symbol $\cdot$ is used as a simplified representation of the sequential tensor-vector multiplication in Eq.~\eqref{eq:TPRM}.
In this case, the current state $\mathbf{h}^{(t)}$ is updated by a system of degree-$p$ homogeneous polynomials of the concatenation of the input and the historic values of the state.
Its more general extensions that include \emph{inhomogeneous} terms can be trivially obtained by padding additional constants along~$\mathbf{x}^{(t)}$.
Although activation functions such as ``tanh'' or ``ReLU'' are popularly used in the model, in this work we consider the case that the activation functions are ignored in order to simplify the theoretical results.
We claim that in practical tasks \emph{the non-trivial tensor power operation has the capability of providing sufficient non-linearity to the model}.

\noindent\textbf{Examples.}
The model~\eqref{eq:TPRM} would degenerate into the most common linear form when $p=1$.
If setting $p=2$, Eq.~\eqref{eq:TPRM} can fully describe the dynamics of models in the works by~\cite{giles1990higher,sutskever2011generating,wu2016multiplicative}, and has a strong connection to weighted finite automata (WTAs)~\citep{rabusseau2019connecting}.
For the higher-degree case, Eq.~\eqref{eq:TPRM} is strongly related to various tensor recurrent models~\citep{yu2017long,huang2020physics,su2020convolutional}.
It is worth noting that in those methods the non-Markovian model is also applied by considering longer historic hidden states into the recurrent process.
Our empirical results show that additional historic states are not necessary for the long memory property. 

\section{Memory of TP recurrent model}
In this section,  we theoretically investigate the memory mechanism of the TP recurrent model~\eqref{eq:TPRM} from both stochastic and stability perspectives, and focus on proving how the degree determines the memory property of the model.
Note that, in the theoretical analysis, we do not apply tensor decomposition~(TD) to the weights, which is different from the works by~\cite{yu2017long,su2020convolutional}.
TD will be exploited in Sec.~\ref{sec:approach} to develop a degree-learnable model.


\subsection{Perspective from RNPs}


Assume that the output of the TP recurrent model is also obtained by the similar degree-$p$ tensor power form to Eq.~\eqref{eq:TPRM} and there is no exogenous input, \textit{i.e.}, \mbox{$\mathbf{x}^{(t)}=\mathbf{y}^{(t-1)}$}, then Eq.~\eqref{eq:RNP} can be specified as
\begin{equation}
\begin{split}
        \mathbf{s}^{(t)}&=\mathcal{M}\times_1{}\mathbf{s}^{(t-1)}\times_2{}\cdots\times_p\mathbf{s}^{(t-1)}+\mathbf{e}^{(t)}\\
        &=\mathcal{M}\cdot\left(\mathbf{s}^{(t-1)}\right)^{\otimes{}p}+\mathbf{e}^{(t)},\quad{}\forall{}t
\end{split},\label{eq:Tensor_Markov}
\end{equation}
where $\mathbf{s}^{(t)}:=\left(\mathbf{y}^{(t),\top},\mathbf{h}^{(t),\top}\right)^\top\in{}\mathbb{R}^{l+m},\,n=l+m$, $\mathcal{M}\in{}\mathbb{R}^{n^{(p+1)}}$ denotes a degree-$(p+1)$ tensor with the symmetric structure among the first $p$ indices, and $\mathbf{e}^{(t)}:=\left(\varepsilon^{(t),\top},0\right)^\top$.
Below, we refer to the stochastic process generated by the model~\eqref{eq:Tensor_Markov} as \emph{tensor-power recurrent network process~(TP-RNP)}.
Below we investigate how the degree $p$ influences the memory of TP-RNP under Def.~\ref{thm:Def}.
First, we show that TP-RNP has short memory if the spectral norm of $\mathcal{M}$ is upper-bounded.
\begin{Lemma}
\label{thm:Prop1}
Under Assumption~\ref{Thm:Assp}, the tensor-power recurrent network process~(TP-RNP)~\eqref{eq:Tensor_Markov} has short memory under Def.~\ref{thm:Def} if the spectral norm of the tensor $\mathcal{M}$ obeys $\Vert{}\mathcal{M}\Vert_{2}<{}1$.
\end{Lemma}


The claim is a natural corollary from Thm.~\ref{thm:zhao}.
It is implied from Lemma~\ref{thm:Prop1} that TP-RNP has the short memory property if the spectral norm of the tensor $\mathcal{M}$ is sufficiently small.
As pointed out by~\cite{zhao2020rnn}, the condition is often satisfied in practice when $p=1$~(\textit{i.e.}, the linear RNN), because the entries of $\mathcal{M}$ are practically bounded away from one. 
However,
\emph{can we expect that the short memory of TP-RNP~\eqref{eq:Tensor_Markov} would be kept if increasing the degree parameter $p>1$?}
To study this point, we specify that the entries of $\mathcal{M}$ obey the sub-Gaussian distribution, \textit{i.e.},
\begin{Assumption}[\textbf{sub-Gaussian and decoupling}]
The tensor $\mathcal{M}$ is obtained by the average over all $p$ first-$p$-indices-shifted variants of a tensor $\mathcal{A}\in{}\mathbb{R}^{n^{(p+1)}}$, of which each entries $\mathcal{A}_{i_1,i_2,\ldots,i_{p+1}}$ is independent, zero-mean and satisfied $\mathbb{E}\left(e^{t\mathcal{A}_{i_1,i_2,\ldots,i_{p+1}}}\right)\leq{}e^{\sigma^2t^2/2}$.\label{thm:Asump2}
\end{Assumption}
It is easily known that the averaged tensor $\mathcal{M}$ is symmetric among the first $p$ indices.
Moreover, because the sub-Gaussian distribution generates samples concentrating around zero with few outliers, the assumption is empirically reasonable as aforementioned.
Next, we theoretically reveal how the degree impacts the model's memory property.
Assuming that our TP-RNP satisfies Assumption~\ref{Thm:Assp} and~\ref{thm:Asump2}, we show below that the long memory requires a high model degree. Specifically,


\begin{Theorem}[\textbf{Long memory requires a high model degree.}]\label{thm:main}
Under Assumptions~\ref{Thm:Assp} and~\ref{thm:Asump2}, with high probability, if TP-RNP~\eqref{eq:Tensor_Markov} has the long memory under Def.~\ref{thm:Def}, then the following inequality obeys:
\begin{equation}
    p\geq{}\frac{p_0}{2}\left(1+\sqrt{1+\frac{C_1}{n\sigma^2}-\frac{C_2}{n}}\right)-1,\label{eq:pbound}
\end{equation}
where $p_0=\log(3/2)$, and $C_1,\,C_2$ denote two positive constants.
\end{Theorem}
The inequality~\eqref{eq:pbound} is obtained using Lemma~\ref{thm:Asump2} and the results on the non-asymptotic bound of tensor spectral norm given by~\cite{tomioka2014spectral}.
We can see that the degree $p$ is controlled by both the dimension of the model and the distribution of $\mathcal{M}$.
Fixed the dimension $n$, a smaller value of $\sigma^2$ leads to a larger value of the degree $p$ for the long memory.

Thm.~\ref{thm:main} implies that the degree $p$ should be sufficiently large, otherwise the model only has short memory~(see Fig.~1 in the \emph{supplementary material} for the visualization of the bound in~\eqref{eq:pbound}).
Intuitively, the model with a higher-degree would result in an unbounded non-linear transition function.
In this case, the transition plays a role like an \emph{``amplifier''}, such that at each time-step both the hidden states and input would be amplified by the degree $p$.
As the result, it would become uneasy for the network to ``forget'' the information a long time ago, \textit{i.e.}, the long memory effect.

\subsection{Perspective from stability analysis}

The long memory effect does not guarantee the recurrent model is trainable.
It is therefore of importance to state clearly whether an unstable behavior would happen in the TP recurrent model.
In particular, \emph{how does the model degree affect the stability of the model?}

Recall the TP recurrent model~\eqref{eq:TPRM}.
To answer the question, we define its stability following the definition given by~\cite{miller2018stable} for an arbitrary recurrent model.
Specifically,


\begin{Definition}[\textbf{stability of TP recurrent model}]
The model~\eqref{eq:TPRM} is stable if there exist some $\lambda<1$ such that, for any states $\mathbf{h},\,\mathbf{h}'$, and input $\mathbf{x}$,
\begin{equation}
    \left\Vert{}\mathcal{G}\cdot
    \left(
    \begin{pmatrix}
    \mathbf{x}\\
    \mathbf{h}
    \end{pmatrix}^{\otimes{}p}
    -
    \begin{pmatrix}
    \mathbf{x}\\
    \mathbf{h}'
    \end{pmatrix}^{\otimes{}p}
    \right)
    \right\Vert_2
    \leq\lambda\Vert{}\mathbf{h}-\mathbf{h}'\Vert_2.\label{eq:DefStability}
\end{equation}\label{thm:DefStability}
\end{Definition}
As shown in Def.~\ref{thm:DefStability}, the TP recurrent model is stable if Eq.~\eqref{eq:TPRM} is $\lambda$-Lipschitz with $\lambda<1$.
Hence, we next prove that \emph{the TP recurrent model is not $L_p$-Lipschitz if $p>1$.}
Before the main claim, we first give a general form of the Jacobian of Eq.~\eqref{eq:TPRM}:
\begin{Lemma}[\textbf{Jacobian of the model}]
\label{thm:Jacobian3}
For any tensor $\mathcal{G}\in{}\mathbb{R}^{n^{p}\times{}m}$ of degree-$(p+1),\,p>0$, the Jacobian matrix $\frac{\partial\mathbf{h}^{(i)}}{\partial\mathbf{h}^{(i-1)}}$ with respect to Eq.~\eqref{eq:TPRM} is equal to 
\begin{equation}
\begin{split}
        &J\left(\mathbf{h}^{(i-1)};\mathbf{x}^{(i)}\right)\\
        &=\frac{\partial\mathbf{h}^{(i)}}{\partial\mathbf{h}^{(i-1)}}=\sum_{k=1}^{p}\left(\mathcal{G}\cdot
        \begin{pmatrix}
        \mathbf{x}^{(i)}\\
        \mathbf{h}^{(i-1)}
        \end{pmatrix}
        ^{\otimes{}p/\{k\}}
        \right)\times_{p+1}
        \begin{pmatrix}
        \mathbf{0}_l\\
        \mathbf{I}_m
        \end{pmatrix}
\end{split},\label{eq:Jacobian}
\end{equation}
where 
$\mathbf{0}_l\in{}\mathbb{R}^{l\times{}m}$ denotes the matrix filled by zeros, $\mathbf{I}_m\in\mathbb{R}^{m\times{}m}$ is a identity matrix, and
the operator $(\,\cdot\,)^{\otimes{}p/\{k\}}$ denotes the sequential ``tensor-vector'' product along the indices in the ordered set $[p]/\{k\}$.
If $\mathcal{G}$ is symmetric among the first $p$ indices, then Eq.~\eqref{eq:Jacobian} can be simplified as
\begin{equation}
\begin{split}
    &J_{s}(\mathbf{h}^{(i-1)};\mathbf{x}^{(i)})
    =p\left(\mathcal{G}\cdot
        \begin{pmatrix}
        \mathbf{x}^{(t)}\\
        \mathbf{h}^{(t-1)}
        \end{pmatrix}
        ^{\otimes{}(p-1)}
        \right)\times_{p+1}
        \begin{pmatrix}
        \mathbf{0}_l\\
        \mathbf{I}_m
        \end{pmatrix}
\end{split}.
\end{equation}
\end{Lemma}
As shown in Lemma~\ref{thm:Jacobian3}, the Jacobian is a constant matrix when $p=1$.
It implies that a linear recurrent model is stable if the spectral norm of $\mathcal{G}$ is sufficiently small.
In this case, the model has the short memory as Lemma~\ref{thm:Prop1}.
Next, we give the main claim by proving the Jacobian~\eqref{eq:Jacobian} is unbounded, which implies the high-degree TP recurrent model would be in an unstable regime.
\begin{Theorem}[\textbf{High degree models lead to unstable behaviors.}]\label{thm:unbound}
Given a tensor $\mathcal{G}$ with the symmetric structure among the first $p$ indices, assume $\mathcal{G}$ has non-zero  sub-blocks with respect to $\mathbf{U}=\left(\mathbf{0}_l^\top\,\,\mathbf{I}_m\right)^\top$, i.e. $\Vert\mathcal{G}\cdot\mathbf{U}^{\otimes{}p}\Vert_2\neq{}0$. 
For any positive number $K>0$ and $p>1$, there always exist a pair of vectors $\mathbf{h}\in\mathbb{R}^{m}$ and $\mathbf{x}\in{}\mathbb{R}^l$, such that $\left\Vert{}J_s(\mathbf{h};\mathbf{x})\right\Vert_2>K$,
i.e., the model~\eqref{eq:TPRM} is unstable under Def.~\ref{thm:DefStability}.
\end{Theorem}
Thm.~\ref{thm:unbound} implies that the TP recurrent model~\eqref{eq:TPRM} is non-Lipschtiz-continuous when $p>1$.
In this case, the model would stay in the unstable regime according to Def.~\ref{thm:DefStability}.
It is known that the gradients of the loss function explode in unstable models~\citep{hardt2018gradient}.
Therefore, Thm.~\ref{thm:unbound} partially reveal the insight why most of (high-degree) tensor models are practically difficult to train, even in the deep feedforward neural networks~\citep{su2020convolutional,huang2020h}.


\subsection{Discussion}
The aforementioned results show a desperate picture:
it seems difficult for the model to obtain the long memory effect meanwhile operating in a stable regime.
To infer its causes, we conjecture \emph{the culprit is the discrete essence of the degree parameter $p$.}
For instance, $p=1$ (\textit{i.e.}, the linear form) provides the model superior stability with a generally short memory; however, the $p=2,3,\ldots$ cases will result in unstable states and a potential long memory.
It inspires us to seek the ``edge'' of such phase transition from the middle of integers~(even less than one).
As the consequence, we could benefit from the both stability and long-term memory in practice.
However, it is non-trivial to achieve the goal because the degree-$p$ tensor power is defined as the multiple folds of tensor products. 
To tackle the issue, in the next section we reformulate the weight tensor $\mathcal{G}$ in Eq.~\eqref{eq:TPRM} using the well-known tensor decomposition~(TD), which allows us to extend the feasible region of $p$ from the intrinsically discrete to a continuous domain.
The extension gives the model the capability of spontaneously learning the optimal degree parameter with respect to the loss function.

\section{Degree-learnable approach}\label{sec:approach}
Below, we show how to extend the discrete degree parameter to a continuous domain, and empirically introduce a degree-learnable approach for the corresponding RNNs.

\subsection{Model description}
Recall the TP recurrent model~\eqref{eq:TPRM} yet adding the bias term for the empirical purpose:
\begin{equation}
    \begin{split}
        \mathbf{h}^{(t)} 
        =\mathcal{G}\cdot
        \begin{pmatrix}
        \mathbf{x}^{(t)}\\
        \mathbf{h}^{(t-1)}
        \end{pmatrix}^{\otimes{}p}
        +\mathbf{b}
    \end{split},\label{eq:TPRMB}
\end{equation}
where $\mathbf{b}\in{}\mathbb{R}^{m}$.
Thus the $j$th entry of the hidden state $\mathbf{h}^{(t)}$, \textit{i.e.} $\mathbf{h}^{(t)}[j]$, can be rewritten as
\begin{equation}
    \mathbf{h}^{(t)}[j]=\left<\mathcal{G}_j,
    \begin{pmatrix}
        \mathbf{x}^{(t)}\\
        \mathbf{h}^{(t-1)}
        \end{pmatrix}^{\otimes{}p}
    \right>
    +\mathbf{b}[j],\label{eq:TPRMBE}
\end{equation}
where $\mathcal{G}_j\in{}\mathbb{R}^{n^p}$ denotes the sub-block of $\mathcal{G}$ by fixing the last index to equal $j$.
Then we know the tensor $\mathcal{G}_j$ is super-symmetric~\citep{cichocki2007nonnegative}.
In this form, the parameter $p$ is only defined in the range of non-negative integers, \textit{i.e.}, $p\in{}\mathbb{Z}^+$.
Next, we apply symmetric tensor decomposition~\citep{comon2008symmetric,brachat2010symmetric} to decomposing the tensor $\mathcal{G}_j$ into factors, \textit{a.k.a.}, latent components~\citep{cichocki2007nonnegative}.
As a result, Eq.~\eqref{eq:TPRMBE} can be represented as 
\begin{equation}
    \mathbf{h}^{(t)}[j]=\sum_{r=1}^R\left<\mathbf{w}_{j,r},
    \begin{pmatrix}
        \mathbf{x}^{(t)}\\
        \mathbf{h}^{(t-1)}
        \end{pmatrix}
    \right>^p
    +\mathbf{b}[j],\label{eq:TPRMBED}
\end{equation}
where $\mathbf{w}_{j,r}\in{}\mathbb{R}^{n}$ denotes the $r$-th factor of $\mathcal{G}_j$ and $R>0$ corresponds to the symmetric tensor rank.
\cite{comon2008symmetric} shows that the decomposition always exists for any symmetric tensor $\mathcal{G}_l$ if the rank $R$ is sufficiently large,
which implies the equivalence between Eq.~\eqref{eq:TPRMBE} and~\eqref{eq:TPRMBED}.
Apart from the equivalence,  we can also see that
\emph{the parameter $p$ is converted into a vanilla exponent, which has explicit definition on not $\mathbb{Z}^+$ but the real field $\mathbb{R}$}.
It allows us to naturally extend the TP recurrent model to the ``real'' degree.
However, given a non-integer $p$, note that the exponential term in Eq.~\eqref{eq:TPRMBED} is defined only when the base~(\textit{i.e.}, the inner product term) is positive.
Therefore, we heuristically extract the sign of the base from exponential function, \textit{i.e.},
\begin{equation}
    \mathbf{h}^{(t)}[j]=\sum_{r=1}^R
    a_{j,r}
    \left\vert\left<\mathbf{w}_{j,r},
    \begin{pmatrix}
        \mathbf{x}^{(t)}\\
        \mathbf{h}^{(t-1)}
        \end{pmatrix}
    \right>\right\vert^p
    +\mathbf{b}[j],\label{eq:TPRMBEDA}
\end{equation}
where $a_{j,r}:=sgn\left(\left<\mathbf{w}_{j,r},
    \begin{pmatrix}
        \mathbf{x}^{(t)}\\
        \mathbf{h}^{(t-1)}
        \end{pmatrix}
    \right>\right)$.
Given any $p\in{}\mathbb{R}$, define an element-wise non-linear function $\phi_p(\cdot):\mathbb{R}^{m}\rightarrow{}\mathbb{R}^{m}$, of which each element $\phi_p(s)$ is given by
\begin{equation}
    \phi_p(s)=sgn(s)\cdot\left\vert{}s\right\vert^p,\label{eq:}
\end{equation}
then we finally have the extended TP recurrent model by concatenating Eq.~\eqref{eq:TPRMBEDA} over all possible $j\in{}[m]$, \textit{i.e.},
\begin{equation}
    \begin{split}
        \mathbf{h}^{(t)}=\sum_{r=1}^R\phi_p\left(
        \mathbf{W}_{hh,r}\mathbf{h}^{(t-1)}+\mathbf{W}_{hx,r}\mathbf{x}^{(t)}
        \right)
        +\mathbf{b}
    \end{split},\label{eq:TPRM_E}
\end{equation}
where $\mathbf{W}_{hh,r}\in\mathbb{R}^{m\times{}m}$ and $\mathbf{W}_{hx,r}\in\mathbb{R}^{m\times{}l}$ are the weights, where the $j$-th row of their concatenation corresponds to the vector $\mathbf{w}_{j,r}$ in Eq.~\eqref{eq:TPRMBEDA}.

It can be seen that the transition of the hidden states in Eq.~\eqref{eq:TPRM_E} results in a \emph{multi-branch} neural network following the activation function $\phi_p$, and the number of branches is determined by the maximum of the symmetric tensor rank of $\mathcal{G}_j,\,\forall{}j\in{}[l]$.

\noindent\textbf{Remark.}
We can see that the degree-induced function $\phi_p$ is able to provide sufficient non-linearity (when $p\neq{}1$) to the learning model even if the standard ``activations'' are omitted.
Moreover, \cite{zhao2020rnn} shows that any saturated continuous activation functions like ``tanh'' and ``sigmoid'' lead to short memory of the model.
Unlike them, the function~$\phi_p$ is unbounded if $p\neq{}0$, and its curvature is controllable in terms of the parameter $p$.
Therefore, the model~\eqref{eq:TPRM_E} can be also considered as an activation-learnable recurrent model, which is expected to have the long-memory effect.

\subsection{Application in RNNs}
The model~\eqref{eq:TPRM_E} can be directly employed to replace the original recurrent model in both vanilla RNN and LSTM.
For the latter,
we suggest a similar way by~\cite{yu2017long}, i.e.,
\begin{equation}
    \begin{split}
        &[\mathbf{i}^{(t)},\mathbf{g}^{(t)},\mathbf{f}^{(t)},\mathbf{o}^{(t)}]
        =\\
        &\sum_{r=1}^R\phi_p\left(
        \mathbf{W}_{hh,r}\mathbf{h}^{(t-1)}+\mathbf{W}_{hx,r}\mathbf{x}^{(t)}\right)+\mathbf{b},\\
        &\mathbf{c}^{(t)}=\mathbf{c}^{(t-1)}\circ{}\mathbf{f}^{(t)},\quad{}\mathbf{h}^{(t)}=\mathbf{c}^{(t)}\circ\mathbf{o}^{(t)}
    \end{split}\label{eq:TPLSTM}
\end{equation}
where $\circ$ denotes the Hadamard product. 
In addition, the trick by considering more historic states as~\citep{soltani2016higher,yu2017long} can be trivially applied to the model~\eqref{eq:TPRM_E}.

Besides the hidden state, we suggest two methods to learning the degree parameter $p$ in the training phase.
The most vanilla way is just to consider $p$ as a trainable variable.
Since its feasible range becomes the whole real field $\mathbb{R}$, $p$ can be efficiently trained by stochastic gradient descent~(SGD) and its variants.
The second way is to learn the parameter $p$ by sub-networks, \textit{i.e.},
\begin{equation}
    p^{(t)}=MLP\left(p^{(t-1)},\mathbf{h}^{(t-1)},\mathbf{x}^{(t)}\right),
\end{equation}
where $MLP(\,\cdot\,)$ denotes a multilayer perceptron~(MLP), and $p^{(t)}$ is the degree parameter at time-step $t$.
Unlike the first method, the sub-network gives the model the variety of the degree parameters for different time-steps.
From the empirical perspective, it is reasonable because the model would have an ``attention-like'' mechanism to selectively remember more important information from the data.




\section{Experiments}
In this section, we numerically show the model~\eqref{eq:TPRM_E} can achieve the superior performance on various time-series forecasting tasks.

\subsection{Single-cell architecture}
We first evaluate the performance of the model on \emph{four long memory datasets}.
To eliminate the irrelevant influence by the network architecture, we compare the models by employing them in a single-cell RNN framework.

\noindent\textbf{Dataset.}
We consider one synthetic dataset and three real-world datasets in this experiment, including ``ARFIMA''~\citep{zhao2020rnn}, ``Dow Jones Industrial Average~(DJI)", ``Traffic''~\citep{UCI2019} and ``tree-ring~(Tree)''~\citep{tsdl}.
The length of the training, validation, and test sets for each dataset is given in the supplementary material.
\cite{zhao2020rnn} shows that the four datasets have strong long-range dependence, \textit{i.e.}, the long memory property.
In the experiment, we perform one-step rolling forecasts on the test set.

\noindent\textbf{Setup.}
In our model, we use a two-layers MLP of the hidden dimension equaling $3$ to calculate the degree parameter $p$, and fix the tensor rank $R$ in Eq.~\eqref{eq:TPRM_E} to equal $1$.
Moreover, we also apply the trick mentioned in~\cite{soltani2016higher,yu2017long} by taking more historic states~(one step more or none) into account, and the model is selected by choosing the one with the best performance on the validation set.
In the training phase, we apply the mean square error~(MSE) as the loss function, and employ the Adam algorithm with learning rate equaling $0.01$ for optimization, which stops after 1000 epochs.

For comparison, we also report the performance of various RNN models given by~\cite{zhao2020rnn}, which includes the vanilla RNN~(RNN), two-lane RNN with the past $100$ values as input~(RNN2), recurrent weighted average network (RWA), MIxed hiSTory RNNs~(MIST), memory-augmented RNN~(MRNN), vanilla LSTM~(LSTM), and memory-augmented LSTM~(MLST).
All the experiments run independently by 50 times, where both the mean value and standard deviation~(std.) of the root-mean-square error~(RMSE) on the test sets are illustrated.

\begin{figure}
    \centering
    \includegraphics[width=1\linewidth]{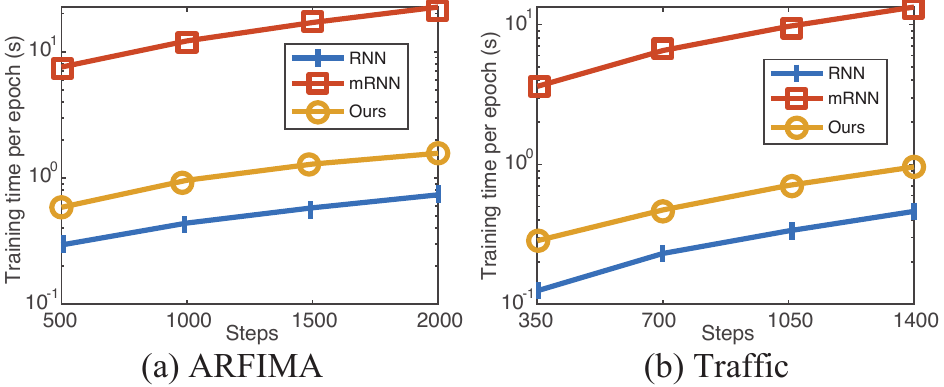}
    \caption{Training time per epoch with various length of the training sets.}
    \label{fig:running_time}
\end{figure}

\noindent\textbf{Results.}
The RMSE performance by various models is shown in Table~\ref{tbl:LT-Performance}.
We can see that our model~\eqref{eq:TPRM_E} outperforms other methods on ``ARFIMA'', ``Traffic'' and ``Tree'', and remains competitive on ``DJI''.
Apart from the mean value, we can also see that our model gives smaller standard deviation than its counterpart like MRNN and MLSTM.
Although RWA gives the smallest std. in the experiment, yet its average performance is not comparable to ours.
In addition, Figure~\ref{fig:running_time} shows the average training time per epoch under various training size.
As shown in~Figure~\ref{fig:running_time}, our model is twice slower than the vanilla RNN yet 10$\times$ faster than MRNN, which has the most closed performance to ours.
\begin{table}
\small
\centering
\caption{Performance comparison in terms of RMSE, where the average and standard deviation~(in brackets) are reported, and the best results are hightlighted in bold.}
\vspace{0.2cm}
\begin{tabular}{lcccc}
\toprule
&ARFIMA&DJI($\times$100)&Traffic&Tree\\
\midrule
\multirow{2}{*}{RNN}&1.1620&0.2605&336.44&0.2871\\
&(0.1980)&(0.0171)&(10.401)&(0.0086)\\
\multirow{2}{*}{RNN2}&1.1630&0.2521&336.32&0.2855\\
&(0.1820)&(0.0112)&(10.182)&(0.0077)\\
\multirow{2}{*}{RWA}&1.6840&0.2689&346.62&0.3048\\
&(0.0050)&(0.0095)&(1.410)&(0.0001)\\
\multirow{2}{*}{MIST}&1.1390&0.2604&358.09&0.2883\\
&(0.1832)&(0.0154)&(16.270)&(0.0091)\\
\multirow{2}{*}{MRNN}&1.0880&\textbf{0.2487}&333.72&0.2818\\
&(0.1140)&(0.0105)&(10.157)&(0.0053)\\
\multirow{2}{*}{LSTM}&1.1340&0.2492&337.60&0.2833\\
&(0.1200)&(0.0128)&(8.146)&(0.0070)\\
\multirow{2}{*}{MLSTM}&1.1490&0.2531&337.83&0.2859\\
&(0.1660)&(0.0130)&(9.440)&(0.0083)\\
\multirow{2}{*}{Ours}&\textbf{1.0691}&0.2672&\textbf{329.22}&\textbf{0.2799}\\
&(0.0245)&(0.0526)&(3.3713)&(0.0023)\\
\bottomrule
\end{tabular}
\label{tbl:LT-Performance}
\end{table}

\vspace{0.5cm}
\noindent\textbf{Larger ranks make the network less non-convex.}
Table~\ref{tab:rank} shows the results of our model on ``ARFIMA'' with various tensor rank $R$.
We can see a counter-intuitive phenomenon that a larger tensor rank would neither improve~(even degrade) the prediction accuracy nor decrease the training loss.
In fact, the similar results have been also reported in several tensor literature~\citep{liu2018efficient,litpfn,huang2020h}.
It is worth noting that, although the averages of loss are almost unchanged when increasing the rank, yet the kurtosis~(a concentration measure of the distribution) significantly increases.
We infer that the model with large ranks generally has a more flat landscape in the training phase, \textit{i.e.}, the loss is less non-convex,
such that the well-trained loss values are less influenced by gradient-based optimizers with different initialization.
This result is empirically consistent with a recent study on the landscape of multi-branch feedforward neural networks~\citep{zhang2019deep}.

\noindent\textbf{Are more historic states really necessary?}
Table~\ref{tab:historic_state} shows the test RMSE of our model on ``ARFIMA'' and ``Tree'' armed with various numbers of historic states as ~\cite{soltani2016higher,yu2017long}, where $D_h=1$ denotes only the current hidden state is used.
It can be seen that additional states would improve the prediction accuracy.
We infer that more historic information of the hidden states would enhance the short-term prediction capability of the model.
On the other hand, the performance would be worse if too much ``history'' is used.
We conjecture that in this case the short-term contribution would \emph{dominate} the prediction performance, such that the model cannot well exploit the long memory effect.

\begin{table}[htp]
\small
    \centering
    \caption{Performance of the proposed model on ``ARFIMA'' under various ranks.}
    \vspace{0.2cm}
    \begin{tabular}{lccccc}
         \toprule
         \multirow{2}{*}{ }&\multicolumn{2}{c}{RMSE}&\multicolumn{3}{c}{Training loss}\\
         &mean&std.&mean&std.&kurtosis\\
         \midrule
         $R=5$&1.0851&0.0270&0.0078&0.0003&2.0620\\
         $R=10$&1.0963&0.0261&0.0079&0.0003&2.6152\\
         $R=30$&1.1057&0.0315&0.0081&0.0004&3.2933\\
         $R=50$&1.1103&0.0316&0.0081&0.0004&3.6301\\
         \bottomrule
    \end{tabular}
    \label{tab:rank}
\end{table}

\begin{table}[htp]
    \small
    \centering
    \caption{Test RMSE of the proposed model on datasets ``ARFIMA'' and ``Tree''. Each row of the table represents the model arms with different numbers of the historic states, where $D_h=1$ denotes only the current hidden sate is used.}
    \vspace{0.2cm}
    \begin{tabular}{lcccc}
         \toprule
         &\multicolumn{2}{c}{ARFIMA}&\multicolumn{2}{c}{Tree}\\
         &mean&std.&mean&std.\\
         \midrule
         $D_h=1$&1.0828&0.0353&\textbf{0.2799}&0.0023\\
         $D_h=2$&\textbf{1.0691}&0.0245&0.2803&0.0021\\
         $D_h=3$&1.0741&0.0388&0.2805&0.0022\\
         $D_h=5$&1.0743&0.0348&0.2803&0.0021\\
         $D_h=10$&1.0835&0.0357&0.2814&0.0018\\
         \bottomrule
    \end{tabular}
    \label{tab:historic_state}
\end{table}



\subsection{Seq2seq architecture}
Next, we evaluate the effectiveness of our model in a deeper architecture, \emph{seq2seq}, on the forecasting task.

\noindent\textbf{Datasets.}
We consider one synthetic dataset and two real-word datasets in the experiment, including ``Genz''~\citep{yu2017long}, ``Traffic of Los Angeles County~(TrafficLA)''\footnote{http://pems.dot.ca.gov} and ``Solar''\footnote{https://www.nrel.gov/grid/solar-power-data.html}.
The preprocessing details and the data format are introduced in the supplementary material.
Unlike the one-step rolling forecast in the single-cell experiment, here we use partial observation to predict the rest of the time series, \textit{i.e.}, relatively long-term prediction.

\noindent\textbf{Setup.}
We follow a similar setup to the experiments given by~\cite{yu2017long}.
In our model, we modify the LSTM cells in the seq2seq network as Eqs.~\eqref{eq:TPLSTM}, and learn the degree parameters by the both trainable variables~(Vanilla) and sub-networks~(Sub-net) as mentioned in Sec. 4.2.
To train the models, we use the sum of square errors as the loss function and RMS-prop as the optimizer.
For comparison, we employ the seq2seq models with both RNN and LSTM in the experiment.
We also compare the performance with the high-order tensor LSTM~(HOTLSTM)~\citep{yu2017long}, which is the most related tensor method to ours.
More details such as the hyper-parameter search are given in the supplementary material.

\begin{figure}
    \centering
    \includegraphics[width=1\linewidth]{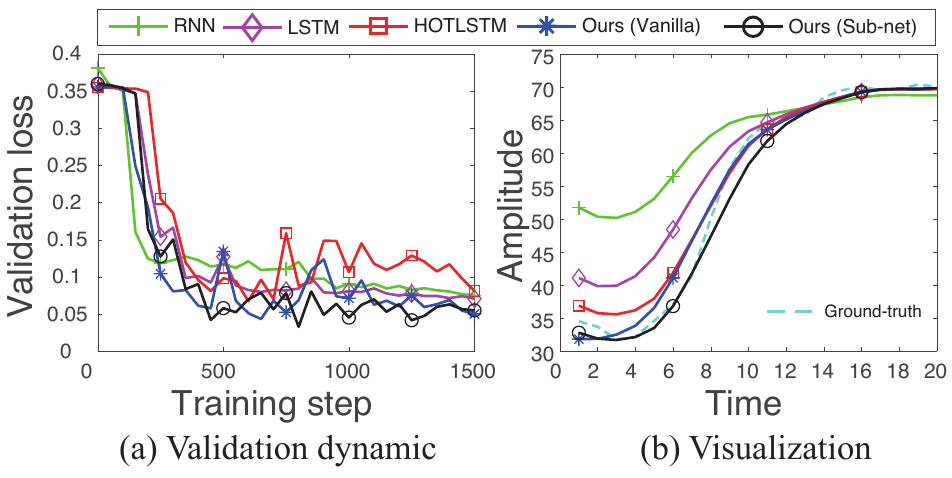}
    \caption{Validation dynamics on ``Genz'' and the visualization of the prediction on ``TrafficLA''.}
    \label{fig:example}
\end{figure}

\begin{table}[htp]
    \small
    \caption{The best RMSE performance on the test sets. In our models, ``vanilla'' indicates directly learning the degree parameters as trainable variables, and ``Sub-nets'' means they are learned by sub-networks.
    The column ``size'' shows the number of trainable parameters for each model~(rounding to 100).}
    \vspace{0.2cm}
    \centering
    \begin{tabular}{lccc||c}
        \toprule
         &Genz&TrafficLA&Solar&Size  \\
         \hline
         RNN&0.0172&0.0874&0.1273&--\\
         LSTM&0.0085&0.0900&0.0947&3300\\
         HOTLSTM&0.0074&0.0851&0.0933&2600\\
         Ours~(Vanilla)&\textbf{0.0061}&0.0853&0.0933&\textbf{2400}\\
         Ours~(Sub-net)&0.0068&\textbf{0.0828}&\textbf{0.0926}&2900\\
         \bottomrule
    \end{tabular}
    \label{tab:Seq2SeqResult}
\end{table}
\noindent\textbf{Results.}
Table~\ref{tab:Seq2SeqResult} shows the best RMSE performance of various models on the three datasets.
We can see that our models outperform not only the conventional RNN and LSTM but also the more advanced tensor-based model HOTLSTM, and the model size of our method is comparable.
We also show the dynamics of validation loss on ``Genz'' and the prediction visualization on ``TrafficLA'' in Figure~\ref{fig:example}.
We can see that our model~(Sub-nets, the black line) consistently obtains superior validation loss with growing the training steps, and prediction results visually demonstrate the effectiveness of our model.


\section{Discussion and concluding remarks}
Our experiments show the recurrent neural networks~(RNNs) can benefit from the \emph{learnable} tensor-power (TP) operations,
and our theoretical results show the degree parameter plays a critical role in the both memory mechanism and dynamic behaviors of the TP recurrent model.

An interesting finding in the experiments is that growing the tensor rank $R$ in Eq.~\eqref{eq:TPRM_E} does not improve the approximation error of the model.
It implies that the rank would not be a key factor to determine the approximation capacity of the model\footnote{Note that it is a different issue from the discussion in the work by~\cite{khrulkov2017expressive}.}.
The result is partially counter-intuitive.
Therefore, the study of the model's approximation theorem remains important for future work.

\section*{Acknowledgement}
We thank the anonymous (meta-)reviewers of \mbox{AISTATS~2021} for helpful comments and discussion.
The work is partially supported by JSPS KAKENHI (Grant No. 20K19875, 20H04249, 20H04208), the National Natural Science Foundation of China (Grant No. 62006045), Ningbo Project (Grant No. NBCP 2019C50052), and Nottingham Project (Grant No. NCHI I01200100023).

\bibliography{references}

\begin{thebibliography}{}

\bibitem[Bengio et~al., 1994]{bengio1994learning}
Bengio, Y., Simard, P., and Frasconi, P. (1994).
\newblock Learning long-term dependencies with gradient descent is difficult.
\newblock {\em IEEE Transactions on Neural Networks}, 5(2):157--166.

\bibitem[Beran et~al., 2016]{beran2016long}
Beran, J., Feng, Y., Ghosh, S., and Kulik, R. (2016).
\newblock {\em Long-Memory Processes.}
\newblock Springer.

\bibitem[Brachat et~al., 2010]{brachat2010symmetric}
Brachat, J., Comon, P., Mourrain, B., and Tsigaridas, E. (2010).
\newblock Symmetric tensor decomposition.
\newblock {\em Linear Algebra and its Applications}, 433(11-12):1851--1872.

\bibitem[Cichocki et~al., 2007]{cichocki2007nonnegative}
Cichocki, A., Zdunek, R., and Amari, S.-i. (2007).
\newblock Nonnegative matrix and tensor factorization [lecture notes].
\newblock {\em IEEE Signal Processing Magazine}, 25(1):142--145.

\bibitem[Comon et~al., 2008]{comon2008symmetric}
Comon, P., Golub, G., Lim, L.-H., and Mourrain, B. (2008).
\newblock Symmetric tensors and symmetric tensor rank.
\newblock {\em SIAM Journal on Matrix Analysis and Applications},
  30(3):1254--1279.

\bibitem[Giles et~al., 1990]{giles1990higher}
Giles, C.~L., Sun, G.-Z., Chen, H.-H., Lee, Y.-C., and Chen, D. (1990).
\newblock Higher order recurrent networks and grammatical inference.
\newblock In {\em Advances in Neural Information Processing Systems}, pages
  380--387.

\bibitem[Greaves-Tunnell and Harchaoui, 2019]{greaves2019statistical}
Greaves-Tunnell, A. and Harchaoui, Z. (2019).
\newblock A statistical investigation of long memory in language and music.
\newblock In {\em International Conference on Machine Learning}, pages
  2394--2403.

\bibitem[Gulcehre et~al., 2014]{gulcehre2014learned}
Gulcehre, C., Cho, K., Pascanu, R., and Bengio, Y. (2014).
\newblock Learned-norm pooling for deep feedforward and recurrent neural
  networks.
\newblock In {\em Joint European Conference on Machine Learning and Knowledge
  Discovery in Databases}, pages 530--546. Springer.

\bibitem[Hardt et~al., 2018]{hardt2018gradient}
Hardt, M., Ma, T., and Recht, B. (2018).
\newblock Gradient descent learns linear dynamical systems.
\newblock {\em The Journal of Machine Learning Research}, 19(1):1025--1068.

\bibitem[Hochreiter and Schmidhuber, 1997]{hochreiter1997long}
Hochreiter, S. and Schmidhuber, J. (1997).
\newblock Long short-term memory.
\newblock {\em Neural Computation}, 9(8):1735--1780.

\bibitem[Hou et~al., 2019]{hou2019deep}
Hou, M., Tang, J., Zhang, J., Kong, W., and Zhao, Q. (2019).
\newblock Deep multimodal multilinear fusion with high-order polynomial
  pooling.
\newblock In {\em Advances in Neural Information Processing Systems}, pages
  12136--12145.

\bibitem[Huang et~al., 2020a]{huang2020physics}
Huang, Y., Tang, Y., Zhuang, H., VanZwieten, J., and Cherubin, L. (2020a).
\newblock Physics-informed tensor-train convlstm for volumetric velocity
  forecasting.
\newblock {\em arXiv preprint arXiv:2008.01798}.

\bibitem[Huang et~al., 2020b]{huang2020h}
Huang, Z., Li, C., Duan, F., and Zhao, Q. (2020b).
\newblock H-owan: Multi-distorted image restoration with tensor 1x1
  convolution.
\newblock {\em arXiv preprint arXiv:2001.10853}.

\bibitem[Khrulkov et~al., 2019]{khrulkov2019generalized}
Khrulkov, V., Hrinchuk, O., and Oseledets, I. (2019).
\newblock Generalized tensor models for recurrent neural networks.
\newblock {\em arXiv preprint arXiv:1901.10801}.

\bibitem[Khrulkov et~al., 2017]{khrulkov2017expressive}
Khrulkov, V., Novikov, A., and Oseledets, I. (2017).
\newblock Expressive power of recurrent neural networks.
\newblock {\em arXiv preprint arXiv:1711.00811}.

\bibitem[Kolda and Bader, 2009]{kolda2009tensor}
Kolda, T.~G. and Bader, B.~W. (2009).
\newblock Tensor decompositions and applications.
\newblock {\em SIAM Review}, 51(3):455--500.

\bibitem[Le et~al., 2019]{le2018learning}
Le, H., Tran, T., and Venkatesh, S. (2019).
\newblock Learning to remember more with less memorization.
\newblock In {\em International Conference on Learning Representations}.

\bibitem[Lechner and Hasani, 2020]{lechner2020learning}
Lechner, M. and Hasani, R. (2020).
\newblock Learning long-term dependencies in irregularly-sampled time series.
\newblock {\em arXiv preprint arXiv:2006.04418}.

\bibitem[Lee et~al., 1986]{lee1986machine}
Lee, Y., Doolen, G., Chen, H., Sun, G., Maxwell, T., and Lee, H. (1986).
\newblock Machine learning using a higher order correlation network.
\newblock Technical report, Los Alamos National Lab., NM (USA); Maryland Univ.,
  College Park (USA).

\bibitem[Li et~al., 2020a]{litpfn}
Li, B., Li, C., Duan, F., Zheng, N., and Zhao, Q. (2020a).
\newblock Tpfn: Applying outer product along time to multimodal sentiment
  analysis fusion on incomplete data.
\newblock In {\em Proceedings of the European Conference on Computer Vision}.

\bibitem[Li et~al., 2020b]{li2020high}
Li, C., Sun, Z., and Zhao, Q. (2020b).
\newblock High-order learning model via fractional tensor network
  decomposition.
\newblock In {\em First Workshop on Quantum Tensor Networks in Machine
  Learning, 34th Conference on Neural Information Processing Systems (NeurIPS
  2020)}.

\bibitem[Liu et~al., 2018]{liu2018efficient}
Liu, Z., Shen, Y., Lakshminarasimhan, V.~B., Liang, P.~P., Zadeh, A.~B., and
  Morency, L.-P. (2018).
\newblock Efficient low-rank multimodal fusion with modality-specific factors.
\newblock In {\em Proceedings of the 56th Annual Meeting of the Association for
  Computational Linguistics (Volume 1: Long Papers)}, pages 2247--2256.

\bibitem[Mehta et~al., 2019]{mehta2019scaling}
Mehta, R., Chakraborty, R., Xiong, Y., and Singh, V. (2019).
\newblock Scaling recurrent models via orthogonal approximations in tensor
  trains.
\newblock In {\em Proceedings of the IEEE International Conference on Computer
  Vision}, pages 10571--10579.

\bibitem[Miller and Hardt, 2018]{miller2018stable}
Miller, J. and Hardt, M. (2018).
\newblock Stable recurrent models.
\newblock In {\em International Conference on Learning Representations}.

\bibitem[Pan et~al., 2019]{pan2019compressing}
Pan, Y., Xu, J., Wang, M., Ye, J., Wang, F., Bai, K., and Xu, Z. (2019).
\newblock Compressing recurrent neural networks with tensor ring for action
  recognition.
\newblock In {\em Proceedings of the AAAI Conference on Artificial
  Intelligence}, volume~33, pages 4683--4690.

\bibitem[Pascanu et~al., 2013]{pascanu2013difficulty}
Pascanu, R., Mikolov, T., and Bengio, Y. (2013).
\newblock On the difficulty of training recurrent neural networks.
\newblock In {\em International conference on machine learning}, pages
  1310--1318.

\bibitem[Pollack, 1991]{pollack1991induction}
Pollack, J.~B. (1991).
\newblock The induction of dynamical recognizers.
\newblock In {\em Connectionist Approaches to Language Learning}, pages
  123--148. Springer.

\bibitem[Rabusseau et~al., 2019]{rabusseau2019connecting}
Rabusseau, G., Li, T., and Precup, D. (2019).
\newblock Connecting weighted automata and recurrent neural networks through
  spectral learning.
\newblock In {\em The 22nd International Conference on Artificial Intelligence
  and Statistics}, pages 1630--1639.

\bibitem[Royston and Altman, 1994]{royston1994regression}
Royston, P. and Altman, D.~G. (1994).
\newblock Regression using fractional polynomials of continuous covariates:
  parsimonious parametric modelling.
\newblock {\em Journal of the Royal Statistical Society: Series C (Applied
  Statistics)}, 43(3):429--453.

\bibitem[Schlag and Schmidhuber, 2018]{schlag2018learning}
Schlag, I. and Schmidhuber, J. (2018).
\newblock Learning to reason with third order tensor products.
\newblock In {\em Advances in Neural Information Processing systems}, pages
  9981--9993.

\bibitem[Soltani and Jiang, 2016]{soltani2016higher}
Soltani, R. and Jiang, H. (2016).
\newblock Higher order recurrent neural networks.
\newblock {\em arXiv preprint arXiv:1605.00064}.

\bibitem[Su et~al., 2020]{su2020convolutional}
Su, J., Byeon, W., Huang, F., Kautz, J., and Anandkumar, A. (2020).
\newblock Convolutional tensor-train lstm for spatio-temporal learning.
\newblock {\em arXiv preprint arXiv:2002.09131}.

\bibitem[Sun et~al., 2018]{sun2018feature}
Sun, Z., Ozay, M., Zhang, Y., Liu, X., and Okatani, T. (2018).
\newblock Feature quantization for defending against distortion of images.
\newblock In {\em Proceedings of the IEEE Conference on Computer Vision and
  Pattern Recognition}, pages 7957--7966.

\bibitem[Sutskever et~al., 2011]{sutskever2011generating}
Sutskever, I., Martens, J., and Hinton, G.~E. (2011).
\newblock Generating text with recurrent neural networks.
\newblock In {\em International Conference on Mahcine Learning}, pages
  1017--1024.

\bibitem[Tjandra et~al., 2017]{tjandra2017compressing}
Tjandra, A., Sakti, S., and Nakamura, S. (2017).
\newblock Compressing recurrent neural network with tensor train.
\newblock In {\em 2017 International Joint Conference on Neural Networks
  (IJCNN)}, pages 4451--4458. IEEE.

\bibitem[Tomioka and Suzuki, 2014]{tomioka2014spectral}
Tomioka, R. and Suzuki, T. (2014).
\newblock Spectral norm of random tensors.
\newblock {\em arXiv preprint arXiv:1407.1870}.

\bibitem[Trinh et~al., 2018]{trinh2018learning}
Trinh, T., Dai, A., Luong, T., and Le, Q. (2018).
\newblock Learning longer-term dependencies in rnns with auxiliary losses.
\newblock In {\em International Conference on Machine Learning}, pages
  4965--4974.

\bibitem[TSDL, 2019]{tsdl}
TSDL (2018 (accessed Jun. 14, 2019)).
\newblock {\em tsdl: Time Series Data Library}.
\newblock https://pkg.yangzhuoranyang.com/tsdl/.

\bibitem[UCI, 2019]{UCI2019}
UCI (2019 (accessed Dec. 28, 2019)).
\newblock {\em Machine Learning Repository metro interstate traffic volumne
  data set}.
\newblock
  https://archive.ics.uci.edu/ml/datasets/\\Metro+Interstate+Traffic+Volume.

\bibitem[Voelker et~al., 2019]{voelker2019legendre}
Voelker, A., Kaji{\'c}, I., and Eliasmith, C. (2019).
\newblock Legendre memory units: Continuous-time representation in recurrent
  neural networks.
\newblock In {\em Advances in Neural Information Processing Systems}, pages
  15570--15579.

\bibitem[Wang and Niepert, 2019]{wang2019state}
Wang, C. and Niepert, M. (2019).
\newblock State-regularized recurrent neural networks.
\newblock {\em arXiv preprint arXiv:1901.08817}.

\bibitem[Wang et~al., 2020]{wang2020kronecker}
Wang, D., Wu, B., Zhao, G., Chen, H., Deng, L., Yan, T., and Li, G. (2020).
\newblock Kronecker cp decomposition with fast multiplication for compressing
  rnns.
\newblock {\em arXiv preprint arXiv:2008.09342}.

\bibitem[Wu et~al., 2016]{wu2016multiplicative}
Wu, Y., Zhang, S., Zhang, Y., Bengio, Y., and Salakhutdinov, R.~R. (2016).
\newblock On multiplicative integration with recurrent neural networks.
\newblock In {\em Advances in Neural Information Processing Systems}, pages
  2856--2864.

\bibitem[Yang et~al., 2017]{yang2017tensor}
Yang, Y., Krompass, D., and Tresp, V. (2017).
\newblock Tensor-train recurrent neural networks for video classification.
\newblock {\em arXiv preprint arXiv:1707.01786}.

\bibitem[Ye et~al., 2018]{ye2018learning}
Ye, J., Wang, L., Li, G., Chen, D., Zhe, S., Chu, X., and Xu, Z. (2018).
\newblock Learning compact recurrent neural networks with block-term tensor
  decomposition.
\newblock In {\em Proceedings of the IEEE Conference on Computer Vision and
  Pattern Recognition}, pages 9378--9387.

\bibitem[Yu et~al., 2017a]{yu2017long}
Yu, R., Zheng, S., Anandkumar, A., and Yue, Y. (2017a).
\newblock Long-term forecasting using higher order tensor rnns.
\newblock {\em arXiv preprint arXiv:1711.00073}.

\bibitem[Yu et~al., 2017b]{yu2017learning}
Yu, R., Zheng, S., and Liu, Y. (2017b).
\newblock Learning chaotic dynamics using tensor recurrent neural networks.
\newblock In {\em ICML Workshop on Deep Structured Prediction}, volume~17.

\bibitem[Zhang et~al., 2019]{zhang2019deep}
Zhang, H., Shao, J., and Salakhutdinov, R. (2019).
\newblock Deep neural networks with multi-branch architectures are
  intrinsically less non-convex.
\newblock In {\em The 22nd International Conference on Artificial Intelligence
  and Statistics}, pages 1099--1109.

\bibitem[Zhao et~al., 2020]{zhao2020rnn}
Zhao, J., Huang, F., Lv, J., Duan, Y., Qin, Z., Li, G., and Tian, G. (2020).
\newblock Do {RNN} and {LSTM} have long memory?
\newblock {\em arXiv preprint arXiv:2006.03860}.

\end{thebibliography}

\end{document}